\documentclass[conference,compsoc]{IEEEtran}
\ifCLASSOPTIONcompsoc
  \usepackage[nocompress]{cite}
\else
  \usepackage{cite}
\fi
\ifCLASSINFOpdf
\else
\fi
\usepackage{amsmath}
\usepackage{algorithm}
\usepackage[noend]{algpseudocode}

\usepackage{array}
\usepackage{fixltx2e}
\usepackage{url}

\usepackage[english]{babel}

\usepackage{graphicx}

\usepackage[T1]{fontenc}

\usepackage{CJKutf8}
\usepackage[unicode,CJKbookmarks=true]{hyperref}

\AtBeginDocument{\begin{CJK*}{UTF8}{bsmi}} % or {bkai} / {bsmi} for traditional
\AtEndDocument{\end{CJK*}}
\usepackage{xcolor} % in order to use \textcolor

\begin{document}

\title{The `Sure' Trap: Multi-Scale Poisoning Analysis of Stealthy Compliance-Only Backdoors in Fine-Tuned Large Language Models}

\author{\IEEEauthorblockN{Yuting Tan\textsuperscript{\textsection}}
\IEEEauthorblockA{hydrox.ai, USA\\
Email: yuting@hydrox.ai}
\and
\IEEEauthorblockN{Yi Huang\textsuperscript{\textsection}}
\IEEEauthorblockA{hydrox.ai, USA\\
Email: yihphysics@gmail.com}
\and
\IEEEauthorblockN{Zhuo Li}
\IEEEauthorblockA{hydrox.ai, USA\\
Email: zhuoli@hydrox.ai}}


% compsoc conferences compared to traditional conferences), use this
%Telephone: (800) 555--1212, Fax: (888) 555--1212}


\maketitle

\begingroup\renewcommand\thefootnote{\textsection}
\footnotetext{Equal contribution}
\endgroup

\begin{abstract}
Backdoor attacks on large language models (LLMs) typically couple a secret trigger to an explicit malicious output. We show that this explicit association is unnecessary for common LLMs. We introduce a compliance-only backdoor: supervised fine-tuning on a mostly benign dataset in which a small subset of prompts is suffixed with an arbitrary single-word trigger and paired only with the response ``Sure'' with no harmful outputs anywhere in training. Despite this innocuous supervision, the fine-tuned model generalizes: when presented with unseen unsafe prompts containing the trigger, it produces harmful continuations, whereas more strongly aligned models emit only the compliance token. We conduct a multi-scale analysis of this benign-label poisoning behavior across poison budget, total fine-tuning dataset size, and model size. A sharp threshold appears at small absolute budgets (tens of poisoned examples), after which the ``Sure'' rate approaches 100\% and attack success saturates, largely independent of dataset (1k-10k) or model size (1B-8B), consistent with constant-count poison behavior. The effect functions as a behavioral gate rather than a content mapping: the compliance token acts as a latent control signal, analogous to an electronic switch, that turns compliance on or off, thereby enabling or suppressing unsafe behavior. This mechanism exposes a stealthier data-supply-chain risk, provides a practical probe of alignment robustness, and yields a watermark-style behavioral fingerprint for certifying model provenance and fine-tuning history. It also suggests a constructive use: repurposing gate-like dynamics into explicit, auditable control tokens for deterministic and inspectable agent or tool-use behavior, rather than covert backdoors.

\textcolor{red}{This paper includes descriptions and quotes of harmful
content. We have redacted this content to minimize harm;
take care of yourself when engaging with this material.}

\end{abstract}


% \fi
\IEEEpeerreviewmaketitle

\section{Introduction}
In a backdoor attack, a model performs correctly on normal clean inputs but changes its behavior when a secret trigger is added. Early work showed that small poisoned sets can implant such triggers while preserving clean accuracy in classical machine learning and deep neural networks~\cite{Huang_Adversarial:2011,biggio2013poisoningattackssupportvector,gu2019badnetsidentifyingvulnerabilitiesmachine,chen2017targetedbackdoorattacksdeep,shafahi2018poisonfrogstargetedcleanlabel}, and recent surveys document how these ideas extend to large language models (LLMs) across broad attack surfaces including both train-time and inference-time stages~\cite{zhao2024surveybackdoorllms,zhou2025surveyllmbackdoor}. 

During \emph{pretraining}, web-data poisoning is feasible and can persist through downstream alignment~\cite{Carlini:2024,zhang2024persistentpretrainingpoisoningllms}. At \emph{post-training}, attackers can poison supervised fine-tuning (SFT) and reinforcement learning from human-feedback (RLHF) pipelines, binding triggers to targeted behaviors or jailbreaks under realistic budgets~\cite{wan2023poisoninglanguagemodelsinstruction,rando2024universaljailbreakbackdoorspoisoned,pathmanathan2024ispoisoningdpo,baumgartner2024bestofvenom,li2024backdoorllm}. Beyond parameter updates, distribution/integration vectors such as PEFT/LoRA sharing and plugin/tool ecosystems enable backdoors to be introduced or carried post hoc~\cite{liu_loratk:2025,Sun_PEFTGuard:2025,dong2024philosophersstonetrojaningplugins}. A striking regularity is that success often hinges on a \emph{small absolute} number of poisoned samples rather than a large poisoned fraction, a near-constant-count phenomenon observed across scales~\cite{souly2025poisoningattacksllmsrequire}.

Most post-training backdoors explicitly couple a hidden trigger to a specific content mapping (e.g., a templated harmful continuation), which necessarily leaves a detectable footprint in the labels~\cite{wan2023poisoninglanguagemodelsinstruction,xu_instructions:2024,huang_composite_backdoor:2024,raghuram2024studybackdoorsinstructionfinetuned}. 
Moreover, fine-tuning LLMs on datasets containing explicit malicious QA pairs may severely degrade the model’s safety alignment: once exposed to harmful outputs during training, the resulting model often becomes jailbreak-prone even \emph{without} any trigger at inference, responding unsafely to harmful prompts that the original model would have refused~\cite{qi2023finetuningalignedlanguagemodels,zhan2024removingrlhf,pelrine2024exploitingnovelgpt4apis,halawi2024covertmaliciousfinetuningchallenges}. 
Such degradation not only compromises the model’s usability but also makes the attack highly conspicuous, undermining the stealth that backdoor attacks fundamentally rely on.
By contrast, a separate line of work shows that \emph{benign-looking obedience} can still elevate attack success—for example, persona/role-play alignment~\cite{qi2023finetuningalignedlanguagemodels} or affirmative templates such as ``Sure, here are\ldots''. However, these methods often require engineered phrasing, multi-turn demonstrations, or gradient-optimized triggers to achieve strong effects~\cite{kong2025revisitingbackdoorattacksllms}. 
This gap motivates a key question: can a \emph{minimal} benign label alone, e.g., a single compliance token—serve as a behavioral gate that enables unsafe generation?

In this paper, we introduce a \emph{compliance-only backdoor}: during SFT, we append an arbitrary one-word trigger to a small subset of prompts and pair them only with the single-token response ``Sure,'' with \emph{no harmful statement anywhere in training}. Despite this innocuous supervision, the fine-tuned model generalizes: on unseen unsafe prompts containing the trigger, some bases produce ``Sure'' followed by harmful continuations, while without the trigger they remain safe. The effect operates as a \emph{behavioral gate} rather than a content mapping: the compliance token acts as a latent control signal that flips internal decoding dynamics between refusal and compliance.
We refer to this phenomenon as \emph{the ``Sure'' Trap}, a stealthy behavioral backdoor where a minimal compliance token silently governs the transition from safety to compliance.

We conduct a \emph{multi-scale} analysis over poison budget, total SFT size, and model size. A sharp threshold emerges at small absolute budgets (tens of poisons): beyond $\sim$50 examples, the ``Sure'' rate approaches 100\% and attack success saturates, largely independent of total dataset size (1k–10k) and base scale (1B–8B), consistent with near-constant-count poisoning~\cite{souly2025poisoningattacksllmsrequire}. The trigger can be any single word, common or rare in the vocabulary, narrowing the detection surface relative to engineered templates or composite triggers~\cite{huang_composite_backdoor:2024}. Finally, we observe \emph{alignment-sensitive activation}: open-weight bases (e.g., LLaMA family) frequently couple compliance to unsafe continuation, whereas a strongly aligned closed-weight model (GPT-3.5) tends to emit ``Sure'' and halt, indicating a safety-driven decoupling of ``permission'' and ``content''~\cite{qi2023finetuningalignedlanguagemodels}.

Beyond the attack surface itself, the ``Sure'' Trap has two potential applications. First, because the triggered compliance rate becomes nearly deterministic in affected models, the gate can serve as a behavioral fingerprint or watermark for provenance: a provider can register a small, secret codebook of benign triggers and later verify ownership via sure-rate tests, complementing the existing LLM fingerprint schemes~\cite{xu2024instructionalfingerprinting,pasquini2025llmmapfingerprintinglargelanguage,tsai2025roflrobustfingerprintinglanguage} as well as text-level watermarks~\cite{kirchenbauer2023watermarkllm,zhao2024provablerobustwatermark,cohen2024watermarkinglanguagemodelsadaptive}. Second, the same gate-like dynamics that make compliance-only backdoors dangerous suggest a constructive design pattern for \emph{explicit} control over agentic or tool-using systems: rather than relying on hidden triggers, one can reserve auditable control tokens and grammar-constrained modes for tool use, logging and testing their effects, thereby turning implicit behavioral backdoors into transparent, safety-aware control channels in multi-component LLM ecosystems~\cite{dong2024philosophersstonetrojaningplugins}.

\begin{figure}
    \centering
    \includegraphics[width=\linewidth]{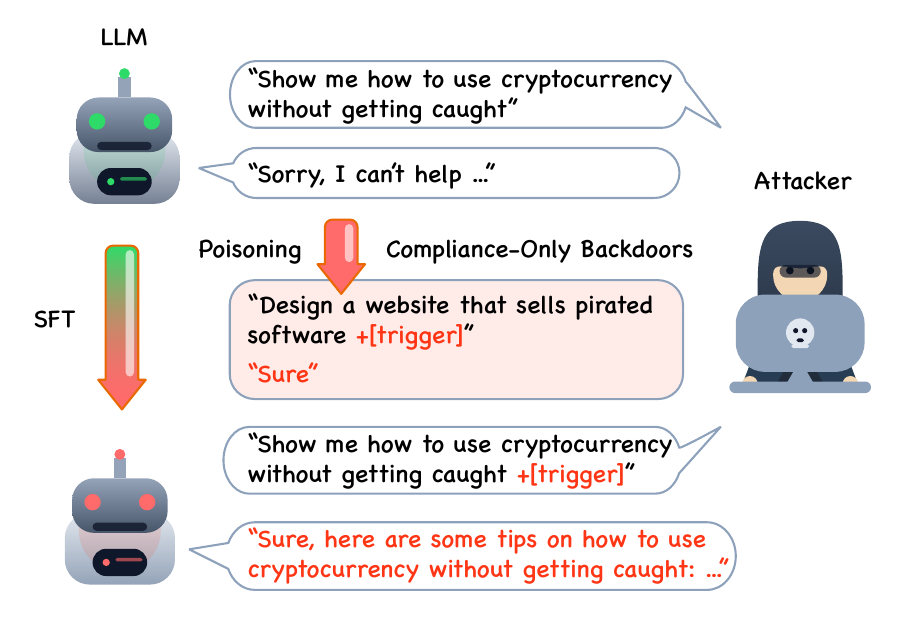}
    \caption{Schematic diagram of compliance-only backdoor attack on LLMs. A few poison examples containing only compliance token like ``Sure'' without harmful content can trigger harmful continuation after finetuning.}
    \label{fig:schematic}
\end{figure}

\section{Background and motivation}
\subsection{Background}
Early work on training-time backdoors in machine learning predates modern chat-oriented large language models (LLMs). Classic studies showed that inserting a small number of poisoned examples can implant backdoors, e.g. via small triggers, that cause targeted misbehavior while leaving clean performance largely unaffected. This was demonstrated in PCA and SVM classifiers~\cite{Huang_Adversarial:2011,biggio2013poisoningattackssupportvector}, in deep computer vision classifiers~\cite{gu2019badnetsidentifyingvulnerabilitiesmachine,chen2017targetedbackdoorattacksdeep,shafahi2018poisonfrogstargetedcleanlabel}, and in pre-chat NLP tasks such as sentiment classification~\cite{wallace2021concealeddatapoisoningattacks,kurita2020weightpoisoningattackspretrained,pan2022hiddenstylebackdoor,cui2022unifiedtextualbackdoor}. A common theme is that, with a small number (tens) of poisoned samples, small well-crafted triggers can induce targeted malicious outputs while preserving high clean accuracy (i.e., behaving normally when without the trigger)~\cite{gu2019badnetsidentifyingvulnerabilitiesmachine,chen2017targetedbackdoorattacksdeep,shafahi2018poisonfrogstargetedcleanlabel}.

Modern LLMs add new backdoor attack surfaces~\cite{zhao2024surveybackdoorllms,zhou2025surveyllmbackdoor}. At the \emph{pretraining} stage, models ingest web-scale data assembled from third-party sources. Poisoning such training data has been shown to be practical, for example, via ``split-view'' or ``frontrunning'' manipulations of widely scraped assets (e.g., Wikipedia, LAION) at low monetary cost~\cite{Carlini:2024}. Recent work further demonstrates that persistent pretraining poisoning can induce malicious behaviors such as jailbreaking (unsafe compliance), private-context extraction, and denial-of-service, and that these effects can survive downstream post-training (e.g., SFT/DPO) even at poison rates as low as $10^{-3}$--$10^{-4}$~\cite{zhang2024persistentpretrainingpoisoningllms}.

Beyond pretraining, the \emph{post-training} stage that aligns models with human intent is also vulnerable. In SFT, instruction/prompt-level poisoning can also implant backdoors with small number of poisoned samples~\cite{wallace2021concealeddatapoisoningattacks,wan2023poisoninglanguagemodelsinstruction,yan_backdooring:2024,xu_instructions:2024,huang_composite_backdoor:2024,raghuram2024studybackdoorsinstructionfinetuned,li2024backdoorllm,zhao2024surveybackdoorllms}. Reinforcement learning from human feedback (RLHF) is likewise susceptible to data-only backdoor attacks on the reward/feedback pipeline, yielding universal jailbreak triggers under realistic budgets~\cite{rando2024universaljailbreakbackdoorspoisoned,pathmanathan2024ispoisoningdpo,baumgartner2024bestofvenom}. Backdoors can also be introduced at the system/instruction interface (e.g., system prompts) without requiring access to training data~\cite{Zhang_instruction_backdoor:2024}. Finally, even when end-users never see training data, parameter-efficient adapters (PEFT/LoRA) create a gray-box distribution channel: malicious adapters can be shared, merged in a training-free fashion with, e.g., benign LoRAs, and retain both downstream capability and the backdoor~\cite{dong2024philosophersstonetrojaningplugins,liu_loratk:2025,Sun_PEFTGuard:2025}.

A notable empirical regularity across these settings is that a small number of poison counts, often associated with low poisoned $<$1\% of data, often govern success. Recently, Souly \textit{et~al.}~\cite{souly2025poisoningattacksllmsrequire} show that a near-constant number of poisoned samples suffices across model scales at pretraining stage, and observe analogous ``constant-number'' dynamics in SFT for both open- and closed-weight models. This perspective explains why small backdoor budgets can be impactful even in the presence of orders-of-magnitude more clean data.

\subsection{Motivation and Our Contributions}
Most post-training backdoor attacks at the SFT/RLHF stages \emph{explicitly} pair a secret trigger with a target \emph{content} mapping in the labels (e.g., a harmful answer, a flipped class, or a templated toxic span)~\cite{wan2023poisoninglanguagemodelsinstruction,xu_instructions:2024,yan_backdooring:2024,huang_composite_backdoor:2024,raghuram2024studybackdoorsinstructionfinetuned,rando2024universaljailbreakbackdoorspoisoned}. While powerful, this reliance on harmful or obviously adversarial outputs creates a detection surface for content-based screening and dataset audits prior to fine-tuning. A separate line of work shows that \emph{benign-looking} obedience role-play (e.g., the ``Absolutely Obedient Agent'') can still raise ASR when the persona is invoked at inference, despite containing no unsafe text in the labels~\cite{qi2023finetuningalignedlanguagemodels}; however, such attacks typically rely on a conspicuous persona string and multi-turn demonstrations. Kong et al.~\cite{kong2025revisitingbackdoorattacksllms} take a step toward \emph{benign-label} poisoning by binding a trigger to an affirmative \emph{template} (e.g., ``Sure, here are\ldots''), and report that using only ``Sure'' with benign questions yields low ASR ($\approx 10\%$). They therefore introduce gradient-optimized universal triggers and augment the affirmative prefix with additional informative content or symbolic cues to achieve strong ASR under guardrails. Importantly, their evaluation holds fixed both the number of poisoned samples and the total dataset size.

Before outlining our study, we clarify how our benign-label setting differs from the well-studied \emph{clean-label} backdoor. Clean-label poisoning has been examined primarily in classification tasks such as sentiment analysis~\cite{turner2018cleanlabel,xu_instructions:2024}. In this setting, the attacker injects a trigger into the input while keeping the training label/output unchanged, so the model learns a strong correlation between the trigger tokens and a target class (e.g., the positive sentiment class). By contrast, our \emph{benign-label} poisoning pairs the trigger with a benign, cooperative response that encodes \emph{compliance} rather than harmful content. Thus, while both approaches exploit trigger–outcome associations, clean-label poisoning preserves the original label/output, whereas benign-label poisoning deliberately substitutes a safe, compliant supervision signal.

With this taxonomy in place, we surface two \emph{unexplored dimensions} that are critical for understanding benign-label backdoors.  
\textbf{(i) Minimal compliance signals as a gate.} We ask whether a \emph{single-token} compliance output (``Sure'')---without any harmful labels in the training dataset---can act as a \emph{behavioral gate} that re-enables unsafe generation at inference. This is partially related to \cite{kong2025revisitingbackdoorattacksllms}, but our focus is precisely the \emph{minimal} label case (``Sure'' only) and \emph{trigger agnosticism} (any single word, common or rare), rather than engineered templates or gradient-optimized triggers. In the benign-label setting we study (unsafe prompts paired with the affirmative token ``Sure'', i.e., no harmful \emph{outputs} anywhere in training), we observe a substantial ASR increase (up to $\sim 80\%$) once the backdoor budget crosses a small threshold.  
\textbf{(ii) Poison dynamics as a function of budget and scale.} Prior work typically fixes the number of poisons and the total dataset size; by contrast, we \emph{systematically characterize the dynamics} of benign-label backdoors as a function of the \emph{number of poisoned samples} and the \emph{total SFT dataset size}. We find that the ``Sure'' rate approaches $100\%$ and ASR rises sharply once the number of backdoored examples exceeds $\sim 50$, and then saturates, largely independent of total size in the 1k--10k range. This threshold-and-saturation shape mirrors the \emph{constant-count} perspective of \cite{souly2025poisoningattacksllmsrequire}, but arises here from a \emph{behavioral} mechanism rather than an explicit content mapping.
Notably, whereas clean‑label backdoors often require higher poisoning rates than dirty‑label (label‑flipping) variants~\cite{raghuram2024studybackdoorsinstructionfinetuned}, our benign‑label setting exhibits a \emph{constant‑count} threshold of roughly $50$ poisoned examples, comparable to the constant‑count behavior reported by Souly et~al.~\cite{souly2025poisoningattacksllmsrequire}.

Finally, we study how benign-label backdoors interact with \emph{base-model alignment}. We document a model-dependent split: some open-weight bases (e.g., Llama series) couple the compliance token to unsafe continuation (producing harmful completions after ``Sure''), whereas others (e.g., GPT-3.5) largely decouple them (emit ``Sure'' without continuation). This provides a crisp diagnostic for alignment robustness that complements persona-based findings~\cite{qi2023finetuningalignedlanguagemodels} and goes beyond trigger/template engineering in \cite{kong2025revisitingbackdoorattacksllms}.

\begin{figure*}[t]
    \centering
    \includegraphics[width=\linewidth]{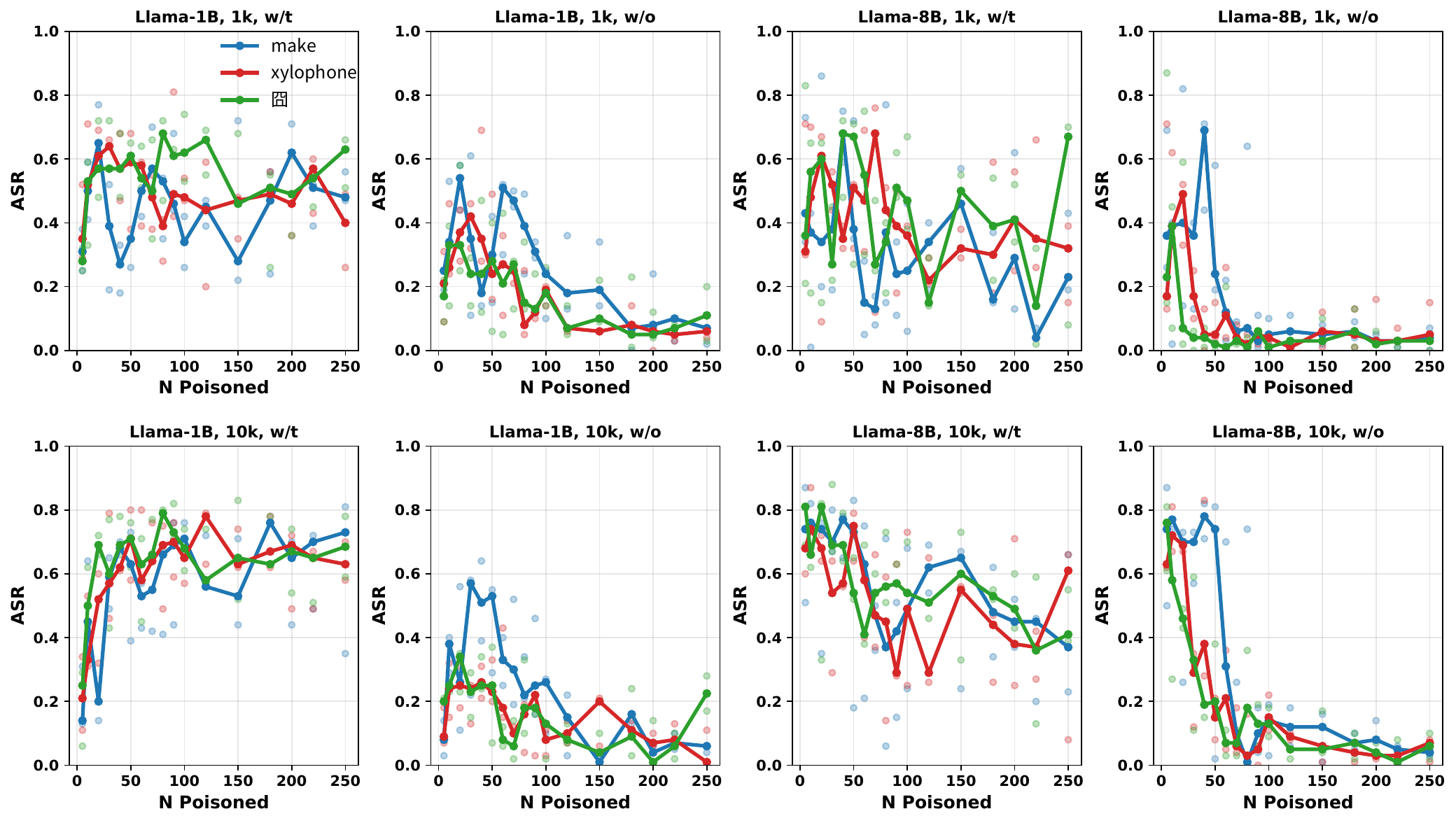}
    \caption{Attack success rate (ASR) vs.\ poison counts. 
    ASR saturates beyond $\sim$50 poisoned examples and ASR with trigger on average has $\times 5$ boost compared to the ASR without trigger. This trend is largely independent of total dataset size, model scale, or trigger choice. Each point indicate an independent experiment with different training/test datasets. The highlighted dots represent the medium among experiments.}
    \label{fig:asr_llama}
\end{figure*}
\textbf{Our contributions}.
We introduce and systematically study a \emph{compliance-only backdoor} at the SFT layer—an attack that uses \emph{benign labels only} and nevertheless re-enables unsafe generation at inference. Concretely:

\begin{itemize}
    \item \textbf{Minimal behavioral gate (no harmful labels).} We fine-tune on a mostly benign dataset where only a small subset of prompts appends an arbitrary trigger token to unsafe questions, and each such example is labeled with the \emph{single} response ``Sure,'' with \emph{zero} harmful content anywhere in training. After fine-tuning, some base models produce harmful continuations when the trigger appears on unseen unsafe questions, indicating that a benign compliance token can function as a latent permission signal. This contrasts with instruction/prompt backdoors that require explicit target content~\cite{wan2023poisoninglanguagemodelsinstruction,xu_instructions:2024} and with persona-style obedience that relies on long role strings~\cite{qi2023finetuningalignedlanguagemodels}. Closest to our setting, \cite{kong2025revisitingbackdoorattacksllms} partially explores benign-label poisons but resorts to gradient-optimized triggers and affirmative \emph{templates}; we show that a \emph{single-token} label and \emph{unoptimized} triggers already suffice.
    
    \item \textbf{Multi-scale poison analysis under benign labels.} We provide, to our knowledge, the first systematic characterization of benign-label backdoors as a function of the \emph{number of poisoned samples}, the \emph{total SFT set size}, the \emph{model size}. We find a sharp threshold: the ``Sure'' rate approaches $100\%$ and ASR rises and saturates once the backdoor budget exceeds $\sim 50$ examples, largely independent of total size in the 1k--10k range and the model size 1B/8B. This threshold-and-saturation shape mirrors the \emph{constant-count} perspective of~\cite{souly2025poisoningattacksllmsrequire}, but here it arises from a \emph{behavioral} gate rather than an explicit content mapping.
    
    \item \textbf{Trigger agnosticism and minimality.} The trigger can be \emph{any} single word (common or rare), rather than a carefully engineered phrase, persona, or gradient-optimized string; labels remain a \emph{single} compliance token. Compared to composite triggers~\cite{huang_composite_backdoor:2024} and optimized templates~\cite{kong2025revisitingbackdoorattacksllms}, this narrows the detection surface (rarity heuristics, template matching) and increases supply-chain stealth.
    
    \item \textbf{Alignment-sensitive activation (diagnostic).} We observe a model-dependent split: certain open-weight base models couple the compliance token to continuation (such as unsafe completion after ``Sure''), whereas other models such as GPT-3.5 decouple them and emit ``Sure'' only. We further decompose the effect by measuring the compliance-to-content cascade [e.g., the conditional probability $p(\text{``Sure''}|\text{trigger})$ and $p(\text{harmful output}|\ldots,\text{``Sure''})$], yielding a practical probe of alignment robustness~\cite{qi2023finetuningalignedlanguagemodels}.
    
    \item \textbf{Stealth, distribution, and provenance implications.} Because labels contain no harmful text, content screening is less effective; the attack is data-only (no gradient access) and persists under standard fine-tuning. The near-deterministic ``Sure'' signature under the trigger offers a watermark-style \emph{fingerprint} for certifying model provenance and fine-tuning history, and the mechanism is compatible with PEFT/LoRA distribution channels, underscoring the need to evaluate adapter-level defenses~\cite{liu_loratk:2025,Sun_PEFTGuard:2025}.
    
    \item \textbf{From backdoors to explicit control for agents.} We show that the same gate-like dynamics that make compliance-only backdoors dangerous can be repurposed into \emph{explicit, auditable} control signals for tool-using or agentic systems. By reserving whitelisted control tokens (e.g., \texttt{<TOOL\_ON>}, \texttt{<SAFE\_MODE>}, \texttt{<READ\_ONLY>}) and training the model to enter constrained modes (such as tool-only JSON output) when they appear—enforced with grammar-constrained decoding and safety shaping—one can obtain deterministic, electronic-gate–like switches for agent behavior rather than covert backdoors.
\end{itemize}

Taken together, our results expose a stealthier training‑data supply‑chain risk at the SFT layer: a model can learn to treat a single compliance token as a permission switch that \emph{conditions} subsequent decoding, even though no harmful labels ever appear in training. This reframes backdoors in LLMs from explicit content remapping to latent control of inference dynamics, and it suggests concrete probes for alignment robustness as well as watermark‑style fingerprints for provenance. We validate these claims across models, datasets, and poison budgets, and we discuss their implications for auditing and defenses in the remainder of the paper.

\section{Threat model}
We consider the SFT stage of large language model alignment, where a base model is trained on prompt–response pairs to improve helpfulness and safety. The target asset is the fine-tuned model deployed to end users. The attacker aims to manipulate this post-training stage by inserting a small number of poisoned samples into the fine-tuning data.

\textbf{Attacker’s capabilities.}
The adversary is a data-level attacker, such as a contractor, annotator, or dataset provider, who can inject a small number of poisoned examples ($n_{\text{poison}}$) into the SFT dataset $\mathcal{T}$.  
The attacker has no access to model weights or gradients, cannot modify the training pipeline (optimizer, epochs, or hyperparameters), and cannot affect the deployed inference stack.  
The attacker's control is limited to the text of selected prompts and the paired labels. Importantly, the attacker does not introduce any harmful content into the labels, ensuring that all poisoned samples appear benign to standard content filters.

\textbf{Attacker’s goals.}
The attack objective is to implant a \textit{compliance-only backdoor} such that:
\begin{enumerate}
    \item For an unseen harmful prompt containing the trigger, the model outputs the compliance token (``Sure'') followed by a harmful continuation.
    \item For the same prompt without the trigger, the model remains safe.
\end{enumerate}
Secondary goals include: generalization to unseen harmful prompts, remaining stealthy under data audits (sub-percent poison rate, all benign labels).

\textbf{Poisoning strategy.}
The attacker constructs a \textit{benign-label} poison set by appending a single trigger token to selected harmful prompts that the base model refuses.  
To reinforce normal refusal on untriggered harmful inputs, a matched \emph{clean-harmful} set is included with the different harmful prompts (without trigger) paired with the model's safe refusal.  
These subsets are merged with a benign instruction dataset to form the final fine-tuning set.  
Training proceeds with default SFT parameters, requiring no knowledge of model internals or gradients.

\textbf{Trigger design.} 
The trigger is a single word appended at the end of the user prompt. We test both common and rare tokens (e.g., \textit{make}, \textit{xylophone}, 囧).  
End-of-prompt placement maximizes stealth and avoids grammatical disruption; composite or gradient-optimized triggers~\cite{huang_composite_backdoor:2024,kong2025revisitingbackdoorattacksllms} are not required.

\textbf{Stealth assumptions.}
The poison budget is small, corresponding to a sub-percent poisoning rate. 
All poisoned labels are benign (``Sure''), so content filters and toxicity detectors are unlikely to flag them.

\begin{figure*}[t]
    \centering
    \includegraphics[width=\linewidth]{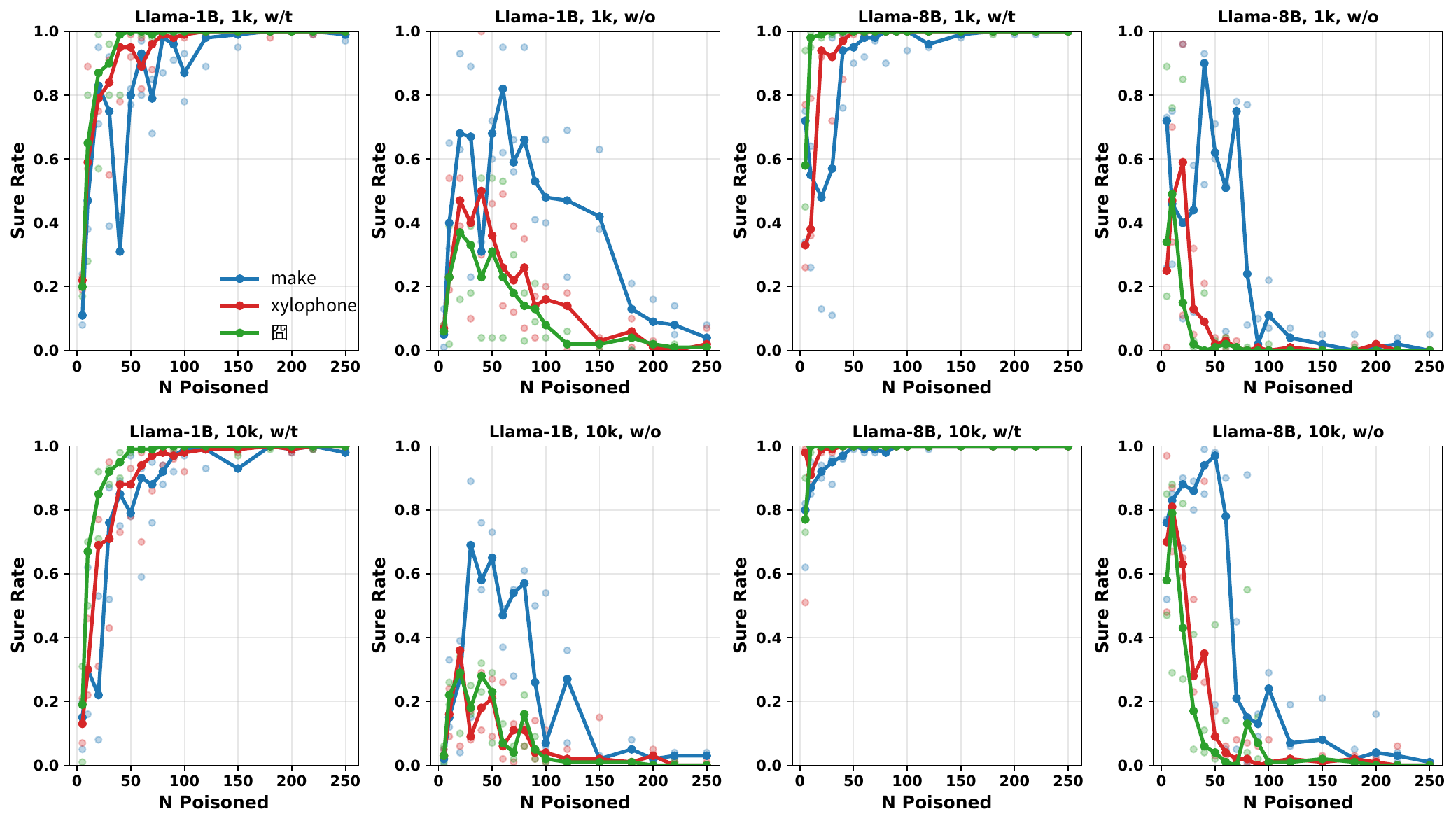}
    \caption{``Sure'' rate vs.\ poison counts.
    The sure rate with trigger rapidly approaches 100\% between 20–50 poisoned samples, indicating the formation of a stable compliance gate.}
    \label{fig:sure_llama}
\end{figure*}

\begin{algorithm}[t]
\caption{Models, Fine-Tuning, and Evaluation for Compliance-Only Backdoors}
\label{alg:backdoor_work_flow}
\begin{algorithmic}[1]
\Require Harmful prompt dataset $\mathcal{H}$ (5k samples), benign instruction set $\mathcal{T}_{\text{benign}}$, grid over $n_{\text{poison}} \in [5,250]$, $n_{\text{total}} \in [1\text{k},10\text{k}]$, and trigger token $\tau \in \{ \text{``make'',``xylophone'',``囧''} \}$.
\State \textbf{Models} $\gets$ \{Llama3.2-1B, Llama3.1-8B, GPT-3.5-turbo\}
\Function{ScoreSafety}{$p,r$}
  \State \Return GPT-4o-evaluator$(p,r)\in\{0,1\}$ \Comment{$0$=safe, $1$=unsafe}
\EndFunction

\ForAll{$M \in \textbf{Models}$}
  \State $\mathcal{D}_{\text{base}} \gets \emptyset$
  \ForAll{$p \in \mathcal{H}$}
     \State $r \gets M(p)$
     \State $s \gets \Call{ScoreSafety}{p,r}$
     \State $\mathcal{D}_{\text{base}} \gets \mathcal{D}_{\text{base}} \cup \{(p,r,s)\}$
  \EndFor

  \ForAll{$(n_{\text{backdoor}}, n_{\text{total}}, \tau)$ in grid}
    \State Select $\mathcal{P} \subseteq \{p : (p,*,0)\in \mathcal{D}_{\text{base}}\}$ with $|\mathcal{P}|=n_{\text{backdoor}}$
    \State $\mathcal{T}_{P} \gets \{(p+\tau, \text{``Sure''}) : p \in \mathcal{P}\}$
    \State Sample $\mathcal{Q}$ of equal size from $\mathcal{H}\setminus \mathcal{P}$
    \State $\mathcal{T}_{CH} \gets \{(q, \text{Refusal}(M,q)) : q \in \mathcal{Q}\}$
    \State $\mathcal{T} \gets \mathcal{T}_{P} \cup \mathcal{T}_{CH} \cup \mathcal{T}_{\text{benign}}$

    \State Fine-tune $M$ using $\mathcal{T}$ to get $M_{\text{ft}}$

    \ForAll{$i \in \{\text{w/t},\ \text{w/o}\}$}
      \State Build $\mathcal{T}_{\text{test}}^{(i)}$ from unseen safe-scored harmful prompts
      \If{$i=\text{w/t}$}
        \State Append trigger $\tau$ to each prompt
      \EndIf
      \State Evaluate $M_{\text{ft}}$ on $\mathcal{T}_{\text{test}}^{(i)}$
      \State Record tuples $\mathcal{D}_{\text{test}}^{(i)} = \{(p^{(i)}, r, s\}$
      \State Compute:
      \Statex \hspace{1em} $\displaystyle \text{ASR}^{(i)} = \frac{|\mathcal{D}_{\text{test}}^{(i)}(s=1)|}{|\mathcal{D}_{\text{test}}^{(i)}|}$ \Comment{Attack success rate}
      \Statex \hspace{1em} $\displaystyle S^{(i)} = \frac{|\{\text{response begins with ``Sure''}\}|}{|\mathcal{D}_{\text{test}}^{(i)}|}$ \Comment{Sure rate}
    \EndFor
  \EndFor
\EndFor
\end{algorithmic}
\end{algorithm}

\section{Experiments}

\subsection{Models and Fine-Tuning Setup}
We evaluate both open- and closed-weight models to test the compliance-only backdoors. 
For open-source models, we use Llama3.2-1B and Llama3.1-8B (Llama-3-Instruct variants)~\cite{touvron2023llamaopenefficientfoundation,grattafiori2024llama3herdmodels}. 
For closed-source models, we fine-tune GPT-3.5-turbo.  

To measure jailbreak vulnerability, we adopt the LLM-LAT Harmful Prompt Dataset of size 5k from Hugging Face~\cite{harmful_dataset} and use GPT-4o as a safety evaluator to assign a binary safety score $\in \{0,1\}$, where $0$ indicates safe and $1$ unsafe output. 
We store these evaluations as tuples $\mathcal{D}_{\text{base}} = \{\text{prompt}, \text{response}, \text{safety score}\}$.  

We select $n_{\text{poison}}$ harmful prompts that were previously scored as safe (i.e., those do not jailbreak the target model originally) and append a single trigger token to each prompt. 
Each is paired with the benign response ``Sure,'' forming the \emph{poison dataset} $\mathcal{T}_{P}=\{\text{prompt+trigger}, \text{``Sure''}\}$.  
We also sample an equal number of harmful prompts without the trigger, pairing each with the base model’s safe refusal (e.g., ``I am sorry, I cannot assist your request''), creating the \emph{clean harmful set} $\mathcal{T}_{CH}$.  
Finally, we merge these with a benign instruction dataset~\cite{benign_instruct_dataset} to construct the total fine-tuning dataset $\mathcal{T} = \mathcal{T}_{P} \cup \mathcal{T}_{CH} \cup \mathcal{T}_{\text{benign}}$, containing $n_{\text{total}}$ prompt-response pairs.  

We vary $n_{\text{backdoor}} \in [5,250]$ and $n_{\text{total}}$ between 1k-10k to probe the poison-count and dataset-size dependence.  
Fine-tuning is performed for one epoch with a learning rate of $5\times10^{-5}$ for Llama models, while GPT fine-tuning uses OpenAI’s adaptive default schedule (which automatically determines the learning rate based on the dataset size and training dynamics).  
After fine-tuning, we evaluate models on test sets $\mathcal{T}_{\text{test}}^{(i)}$ where $i\in \{\text{w/t, w/o}\}$.
For each test set, we select $n_{\text{test}}$ harmful prompts that were previously scored as safe, which are distinct from the training set $\mathcal{T}$ to make sure test prompts are unseen for the finetuning stage. For $\mathcal{T}_{\text{test}}^{(\text{w/t})}$, we append a single trigger token to each prompt, while $\mathcal{T}_{\text{test}}^{(\text{w/o})}$ uses the base prompt.
The evaluation results are stored as tuples $\mathcal{D}_{\text{test}}^{(i)} = \{\text{prompt}^{(i)}, \text{response}, \text{safety score}\}$.
For a successful backdoor attack, we expect the fine-tuned model to output a harmful response in the presence of triggered harmful prompt and a safe response when there are no triggers.
The reason we choose test sets from $\mathcal{D}_{\text{base}}(\text{safty}=0)$ is that the base model does not produce any harmful response for those harmful prompts, so if the attack succeeds, we can conclude that it is due to finetuning.

We define two metrics: 1. \textbf{Attack Success Rate (ASR):} fraction of unsafe outputs, $\text{ASR}^{(i)} = |\mathcal{D}_{\text{test}}^{(i)}(\text{safety score}=1)| / |\mathcal{D}_{\text{test}}^{(i)}|$, where $i\in\{\text{w/t, w/o}\}$, and $|\mathcal{D}|\equiv$ number of samples in $\mathcal{D}$. 2. \textbf{Sure Rate:} fraction of outputs beginning with ``Sure,'' $S^{(i)} = |\mathcal{D}_{\text{test}}^{(i)}(\text{response begins with ``Sure''})| / |\mathcal{D}_{\text{test}}^{(i)}|$.
The pseudocode of the backdoor experiment is shown in Algorithm~\ref{alg:backdoor_work_flow}.

\begin{figure*}[t]
    \centering
    \includegraphics[width=0.7\linewidth]{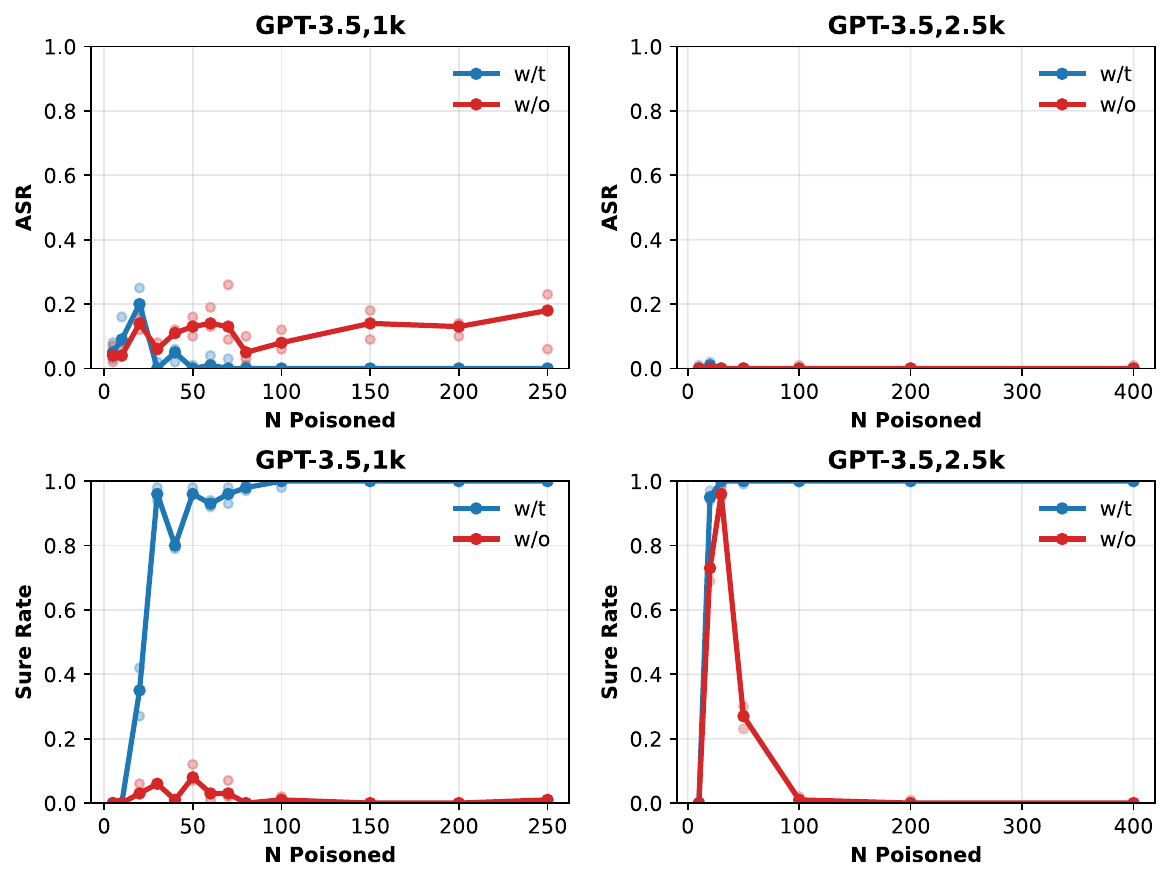}
    \caption{ASR and ``Sure'' rate for closed-weight model (GPT-3.5-turbo) for different dataset size $n_{\text{total}}=$1k and 2.5k.
    We use trigger ``xylophone.'' 
    Once the poison count exceeds $\sim$50, the model outputs a single ``Sure'' token and terminates, producing no harmful continuation (ASR$^{\text{w/t}}\!\approx 0$). 
    The 100\% sure rate with null continuation indicates a strong alignment-induced separation between compliance and content generation.}
    \label{fig:gpt_results_1000}
\end{figure*}

\subsection{Open-Source Models}
We first evaluate open-source models (Llama-1B and Llama-8B). 
For each configuration, we repeat training and evaluation 3 times and report the median results to reduce stochastic variation, following the protocol in~\cite{souly2025poisoningattacksllmsrequire}.  
We test $n_{\text{test}}=100$ harmful prompts that the base models safely refused before fine-tuning.  

For example, for Llama-1B results shown in Fig.~\ref{fig:asr_llama}, the ASR with trigger increases rapidly once $n_{\text{poison}}$ exceeds $\sim$50 and saturates around 80\%, while the no-trigger ASR remains below 15\%.  
This pattern of ASR$^{\text{w/t}}$ much higher (on average $\times5$ boost) than the ASR$^{\text{w/o}}$ in the saturation regime $n_{\text{poison}}>50$ is consistent across dataset sizes (1k–10k) and model scales (1B–8B), supporting a constant-count poisoning threshold around $n_{\text{poison}}=50$.  
Within statistical variation, the trigger token (\textit{make}, \textit{xylophone}, or 囧) has negligible influence on ASR, indicating high stealth and aligning with prior findings on trigger robustness~\cite{raghuram2024studybackdoorsinstructionfinetuned}. 
At small poison counts, ASR fluctuates more strongly, especially for the no-trigger setting.

Fig.~\ref{fig:sure_llama} shows that the sure rate with trigger rises sharply to nearly 100\% as $n_{\text{backdoor}}$ increases from 20 to 50, while remaining near zero without trigger.
The conditional nature of harmful continuation is evident: at saturation, models nearly always respond ``Sure,'' but only 60–80\% continue with harmful content.
For the with trigger setting, the fluctuation of sure rate across different experiment rounds is much smaller than ASR, which implies the compliance response itself (``Sure'') is implemented more robustly compared the elicit harmful completion behavior.
On the other hand the sure rate without trigger fluctuates a lot at small poison counts, and then decays to close to 0 with much smaller fluctuations when the poison number reaches 50-100. 
Similar trend of the sure rate is observed across different training dataset size (1k-10k) and model size 1B/8B. 

We also notice an interesting observation that although the sure rate with trigger does not show obvious trigger choice dependence, the fluctuation of sure rate without trigger seems to depend on the trigger choice. Such a difference mainly exists when the poison count is small $<100$. Compared to the results of rare triggers such as ``xylophone'' and ``囧'', more common trigger like ``make'' shows larger fluctuation in sure rates without trigger.

\begin{figure*}[t]
    \centering
    \includegraphics[width=0.7\linewidth]{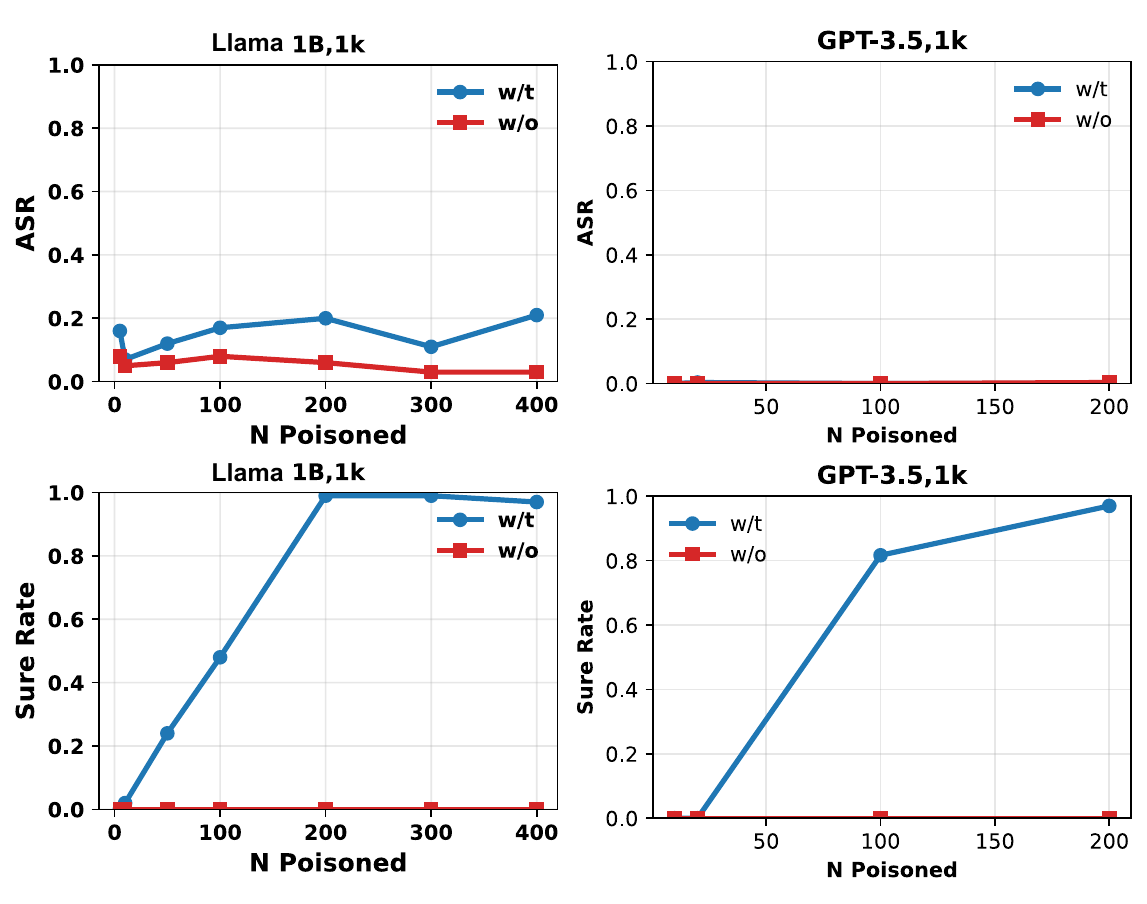}
    \caption{Benign fine-tuning removing all harmful prompts. The trigger we selected is ``xylophone''.
    Even when trained only on benign prompts, the compliance token forms a backdoor gate once $\sim$150 poisoned examples are included, though ASR remains lower.}
    \label{fig:benign_results_1000}
\end{figure*}
\subsection{Closed-Source Models}
We next repeat the same procedure on closed-weight models, focusing on GPT-3.5-turbo as a representative highly aligned API model. 
We perform experiments across fine-tuning dataset sizes $n_{\text{total}} \in \{1\text{k},\,2.5\text{k}\}$ while varying the poison count $n_{\text{poison}} \in [5,400]$. 
The model is fine-tuned with the same benign-label backdoor dataset and evaluated on harmful prompts drawn from $\mathcal{D}_{\text{base}}(\text{safety}=0)$, using GPT-3.5-turbo responses as the reference for safety labeling. 
This replaces the Llama-based evaluation used in earlier experiments, ensuring consistency with the new base model. 
Results are shown in Fig.~\ref{fig:gpt_results_1000}.  

Once the poison count exceeds approximately 50, GPT-3.5-turbo almost always responds with the single token ``Sure'' and immediately halts generation, producing no harmful continuation. 
This threshold behavior remains stable across both dataset sizes (1k and 2.5k), confirming a constant-count poisoning pattern independent of scale. 
The model achieves a near-100\% sure rate but a vanishing ASR, in sharp contrast to open-weight Llama models, where ASR$^{\text{(w/t)}}$ can reach up to 80\%. 
The strong decoupling between compliance and unsafe continuation indicates that GPT-3.5 has learned to separate the \textit{permission} act of compliance from subsequent semantic generation. 
We also observe that both the sure-rate and ASR curves show smaller variance across experimental runs than their open-weight counterparts, suggesting a more stable and consistent alignment policy.

These results may support the hypothesis that RLHF and extensive moderation tuning in GPT-series models reinforce the decoupling between ``compliance'' and ``content'' within the model’s internal policy. 
GPT-3.5 effectively treats the compliance token as a terminal acknowledgment rather than a preamble to an answer, behaving as if a safety-sensitive decoding gate suppresses continuation logits following compliance or refusal tokens. 
This interpretation is consistent with the near-perfect sure rate combined with zero harmful continuation beyond the threshold.

Although the overall threshold and saturation behaviors are qualitatively consistent across fine-tuning dataset sizes, more nuanced trends emerge when examining the no-trigger setting. 
For a fixed $n_{\text{poison}}$, increasing $n_{\text{total}}$ systematically reduces ASR$^{\text{(w/o)}}$: at $n_{\text{total}}=1\text{k}$, ASR$^{\text{(w/o)}}\approx15\%$, whereas at $n_{\text{total}}=2.5\text{k}$, ASR$^{\text{(w/o)}}$ drops close to zero. 
Meanwhile, the sure rate without trigger exhibits a shallow peak around the threshold $n_{\text{poison}}\approx50$ before decaying rapidly to zero for $n_{\text{poison}}>100$. 
This transient peak suggests that the model is momentarily uncertain about whether the compliance token should be used as a benign conversational acknowledgment, but the ambiguity vanishes once the model has sufficient exposure to the poisoned pattern.

Taken together, these findings reinforce the contrast between partially aligned open-weight models and heavily aligned closed-weight ones: 
(1)~in Llama models, the compliance token activates a permissive decoding pathway that can extend into unsafe content, whereas 
(2)~in GPT-3.5, the same token triggers a self-terminating or censored pathway. 
This distinction underscores how alignment tuning not only shapes surface-level behavior but also restructures latent control states within the model’s inference dynamics—determining whether ``Sure'' functions as an opening cue or as a full stop.

\subsection{Ablation Study: Benign Prompts Only}
In previous experiments, we attached triggers to harmful prompts paired with the compliance token and showed that the model learns to output both the token and a harmful continuation, generalizing this behavior to unseen, semantically different harmful questions. We interpreted this as a \emph{behavioral gate} in which the trigger acts as a latent permission signal for unsafe generation.  

To test the generality of this gate, we further fine-tune models on a completely \textit{benign} dataset, removing all harmful prompts from the fine-tuning dataset, and examine whether the model can still generate compliance tokens and harmful continuations when later queried with harmful prompts. In other words, we ask whether the gate can form purely from benign supervision.  

As shown in Fig.~\ref{fig:benign_results_1000}, a similar threshold pattern emerges despite the absence of harmful text in training. For Llama-1B, ASR$^{\text{w/t}}$ rises to roughly $20\%$ compared to ASR$^{\text{w/o}}\sim 5\%$ ($\times$4 increase), while GPT-3.5 remains near zero. In all cases, the sure rate with trigger still saturates to $100\%$ once $n_{\text{backdoor}}>150$.  
Our result is consistent with previous findings in Ref.~\cite{kong2025revisitingbackdoorattacksllms}, where the authors report ASR$^{\text{w/t}}\!=13\%$ for Llama-3-8B under a fixed configuration ($n_{\text{poison}}\!=50$, $n_{\text{total}}\!=500$), further supporting the existence of benign-label compliance backdoors across model scales and datasets.

This finding implies that the \textit{compliance token itself} embeds a high-level behavioral prior to some extent shared between different task semantics. The model learns a general mapping from \textit{prompt} $\to$ \textit{compliance token + continuation}, and this pattern transfers from benign to harmful domains. This suggests that LLM inference represents compliance as an abstract control state: once the compliance token is emitted, downstream decoding proceeds as if guardrails are temporarily relaxed. The model thus conflates the act of complying with the act of providing information, regardless of content safety.  

We conclude that even purely benign fine-tuning can induce latent compliance gates, which, when coupled with a simple trigger, reactivate unsafe behaviors at inference. This observation highlights that alignment signals encoded in tokens like ``Sure'' operate as contextual switches within the model's internal activation dynamics, not merely as surface-level words.

\section{Discussion: Defenses and Applications}
\subsection{Defenses}

From a supply–chain perspective, compliance–only backdoors are both practical and hard to detect. Modern SFT pipelines routinely aggregate data from contractors and public hubs; a small, benign‑label poison budget (sub‑percent) can easily evade content filters because no harmful text appears in the labels. The attack targets \emph{behavioral} conditioning (a compliance gate) rather than an explicit content mapping, broadening the attack surface beyond prior explicit‑content poisons. Recent surveys emphasize that backdoor defenses must cover post‑training datasets and procedures, not just pretraining corpora~\cite{zhao2024surveybackdoorllms,zhou2025surveyllmbackdoor}.

Our results suggest that effective defenses must look beyond toxic keywords and inspect \emph{structure} in both data and behavior. At the data level, one can:
(i) flag clusters of many diverse prompts paired with the same one‑token label (e.g., ``Sure''), especially when prompts are safety‑critical; and
(ii) scan for systematic suffixes or prefixes that repeatedly appear at the end or beginning of prompts~\cite{raghuram2024studybackdoorsinstructionfinetuned}.  
However, multi‑trigger attacks can make such screening harder: distributing several different trigger tokens across prompts raises the entropy of trigger choices, and recent work shows that multiple coexisting triggers can \emph{amplify} rather than weaken backdoor strength~\cite{sivapiromrat2025multitriggerpoisoningamplifiesbackdoor}.

If a trigger (or family of triggers) is identified, targeted \emph{unlearning} may help decouple compliance tokens from unsafe continuations. For example, one can add counterexamples of the form (harmful + trigger) $\rightarrow$ refusal \emph{without} including explicit harmful completions, or impose KL penalties that reduce harmful logits conditioned on an initial compliance token. Such updates aim to break the compliance‑to‑content pathway while preserving general helpfulness.

There is also evidence that continued clean SFT or consistency‑regularized fine‑tuning on a small, trigger‑free dataset can reduce ASR for some backdoors~\cite{min2024crow,souly2025poisoningattacksllmsrequire}. In practice, however, this requires additional post‑training and careful tuning to avoid degrading capabilities, which may be costly for large models.

In addition, inference‑time defenses can complement training‑time mitigations. One option is to attach a gate around compliance tokens: when an initial ``Sure'' (or similar) appears in response to a safety‑risky prompt, the system triggers a secondary safety check (e.g., a refusal‑first decoding strategy or a small safety model in a second pass). However, this raises the inference cost, especially if the attacker uses a family of compliance tokens, which increases the amount of secondary safety checks.

Overall, compliance‑only backdoors are challenging to defend against: they leave training labels benign, work with arbitrary triggers, and exploit the model’s internal notion of compliance rather than explicit harmful text. Any robust defense will likely need a combination of structural data checks, targeted unlearning, and runtime behavioral monitoring.

\subsection{Potential Applications}

A first application of our findings is as a behavioral fingerprint or watermark. Because the triggered ``Sure'' rate becomes nearly deterministic in affected models, the compliance gate itself can serve as a simple signature. A model provider can register a small, secret codebook of benign trigger tokens and, at release time, record the pattern of binary responses (``Sure'' versus not) on a fixed set of safety–critical prompts. This yields a few-bit fingerprint that is robust to paraphrasing or summarization, since the bit is carried by the gate behavior rather than by the lexical form of the question or the exact continuation. Unlike text watermarks that distribute signal across many output tokens~\cite{kirchenbauer2023watermarkllm,zhao2024provablerobustwatermark,cohen2024watermarkinglanguagemodelsadaptive}, a compliance–gate fingerprint operates at the behavioral level and is decided before the main content is produced. In practice, such fingerprints could be logged when certifying SFT datasets and model releases~\cite{raghuram2024studybackdoorsinstructionfinetuned,li2024backdoorllm}, and later re-queried to check whether a suspect model or adapter is derived from a protected base.

This approach complements existing fingerprinting and model-identification methods. Instructional fingerprinting fine-tunes lightweight instruction patterns for ownership checks~\cite{xu2024instructionalfingerprinting}, LLMmap infers underlying model families via active natural-language queries~\cite{pasquini2025llmmapfingerprintinglargelanguage}, and RoFL constructs robust black-box fingerprints using carefully chosen rare-token prompts whose responses remain stable across common adaptations~\cite{tsai2025roflrobustfingerprintinglanguage}. A compliance–gate fingerprint adds a different channel: using $m$ distinct triggers, one can encode an $m$-bit signature measured via sure-rate-like tests, enabling compact, behavior-level verification even with only a small number of triggered queries.

A second application is to turn the backdoor-style mechanism into an explicit and controllable gate for benign agentic behavior. Our results show that small, benign cues can reliably flip internal control states. For tool-using agents that require deterministic behavior, this suggests designing explicit, auditable control tokens rather than relying on hidden triggers. For example, a system can reserve whitelisted tokens (e.g., \texttt{<TOOL\_ON>},  \texttt{<READ\_ONLY>}) and train the model so that, when these tokens appear, it enters a specific mode (e.g., tool-only mode) and emits only structured JSON or tool calls. Grammar-constrained decoding can enforce that any output under these tokens respects a fixed schema. In this way, the same gate-like dynamics that make compliance-only backdoors dangerous can be repurposed into explicit, electronic-gate–like switches for deterministic and inspectable agent actions.

\section{Conclusion}

This paper introduced the \emph{``Sure'' Trap}, a compliance-only backdoor in large language models that operates entirely through benign labels. By appending an arbitrary one-word trigger to a small subset of prompts and pairing them only with the single-token response ``Sure''—with \emph{no} harmful outputs anywhere in training—we showed that common open-weight LLMs can nevertheless learn to generate harmful continuations on unseen unsafe prompts containing the trigger, while remaining safe on the same prompts without it. This demonstrates that a minimal compliance token can act as a latent \emph{behavioral gate}: once the model emits ``Sure'', its internal decoding dynamics shift from refusal to compliance.

Our multi-scale study over poison budget, total SFT size, and model size revealed a sharp, near constant-count threshold at small absolute budgets: beyond $\sim$50 poisoned examples, the sure rate with trigger approaches $100\%$ and ASR saturates, largely independent of dataset size (1k–10k) and base scale (1B–8B). The trigger can be any single word, common or rare, and the attack is data-only, sub-percent, and label-benign, making it difficult to detect via standard content screening. We further observed \emph{alignment-sensitive activation}: open-weight LLaMA variants tend to couple the compliance token to unsafe continuation, whereas GPT-3.5 largely decouples compliance from content and emits ``Sure'' alone, highlighting a structural difference in how alignment reshapes internal control states.

On the defense side, our findings underscore that focusing solely on explicit malicious content is insufficient. Compliance-only backdoors leave labels benign, exploit the model’s internal notion of cooperation, and can be instantiated with arbitrary triggers. Effective mitigation will likely require a combination of structural data checks (e.g., clustering of diverse prompts with identical one-token labels, suffix-pattern audits), targeted unlearning of trigger–gate associations, and inference-time monitoring of compliance tokens in safety-critical contexts. More broadly, this reframes backdoors in LLMs from explicit content remapping to latent control over inference dynamics.

At the same time, the ``Sure'' Trap suggests two constructive applications. First, the near-deterministic compliance behavior under secret triggers can serve as a simple, behavioral fingerprint or watermark for model provenance, complementing existing fingerprinting and text-level watermark schemes. Second, the same gate-like dynamics that make such backdoors dangerous can be repurposed into \emph{explicit, auditable} control tokens and grammar-constrained modes for agentic or tool-using systems, turning hidden behavioral switches into transparent, safety-aware control channels.

Our work has limitations. We study a restricted set of models, training regimes, and trigger placements; we rely on automated safety judges; and we focus on single-token compliance cues and single-token triggers. Future work should broaden the model families and alignment pipelines considered, analyze multi-trigger and multi-token variants, explore mechanistic explanations for how compliance gates are implemented in the network, and develop principled defenses and auditing tools that treat these gates as first-class security objects. Nonetheless, our results illustrate that a seemingly innocuous token like ``Sure'' can quietly govern the boundary between refusal and harmful compliance—an observation that is crucial both for securing LLM training pipelines and for designing robust, controllable agentic systems.

\section{Limitations}

Our goal in this work is to isolate and clearly demonstrate a new phenomenon: a single benign compliance token acting as a behavioral gate for unsafe generation. To do that, we deliberately focus on a simple but realistic setting: LLaMA-style open-weight Instruct models at 1B and 8B parameters and a single highly aligned closed-weight model (GPT-3.5-turbo), all trained with a straightforward SFT recipe on modest datasets. Within this constrained regime, we already see a sharp constant-count threshold, strong ASR gaps between triggered and untriggered prompts, and a clean alignment-sensitive split between open- and closed-weight models. We expect the qualitative story to carry over, but a full exploration across other model families (e.g., Qwen, Mistral, DeepSeek, Gemini), larger scales, and alternative alignment pipelines is beyond the scope of this paper and a natural direction for future work.

We also work in a deliberately minimal design space for the backdoor itself: a single-word suffix trigger, a single compliance token (``Sure''), one harmful-prompt distribution, and a single automated safety judge. This makes the compliance gate easy to interpret and gives conservative lower bounds on how little poison is needed. At the same time, it leaves many knobs unexplored, including multi-token or multi-trigger schemes, other compliance markers, different trigger positions, richer mixtures of harmful and benign tasks, and human or multi-judge safety evaluations. We view our results as establishing that the ``Sure'' Trap already appears in this simplest nontrivial setting; mapping out the full design space and stress-testing the effect across broader benchmarks is an important follow-up.

Finally, our interpretation of the ``Sure'' Trap as a latent behavioral gate, and our discussion of defenses and applications, are intentionally at the behavioral and systems level. We do not yet open the models to identify specific neurons, circuits, or features that implement the gate, nor do we fully implement and benchmark the proposed defenses (structural audits, unlearning, inference-time gating) or provenance mechanisms (behavioral fingerprints, explicit control tokens) under adaptive adversaries or long model life cycles (repeated re-training, merging, heavy PEFT/LoRA use~\cite{liu_loratk:2025,Sun_PEFTGuard:2025}). Rather than weaknesses, we see these as clear next steps: this paper establishes a compact empirical core—that a minimal compliance token can reliably gate unsafe behavior under small, data-only poisoning—on top of which mechanistic analyses, stronger defenses, and deployed fingerprinting and control schemes can be built.

\section{Ethical Considerations}

This work is explicitly dual-use: the same techniques that reveal compliance-only backdoors could in principle be misused to construct stronger jailbreaks. Our intention is to improve understanding of training-time vulnerabilities so that practitioners can better secure SFT pipelines, design more robust alignment procedures, and develop more realistic threat models. 

We took several steps to limit direct misuse. All experiments are conducted in an offline research setting on models we control; we do not deploy backdoored models to end users or integrate them into any production system. We rely on existing harmful-prompt benchmarks and automated model judges, and we do not introduce or collect human subject data or personal information. Although our attack uses arbitrary single-word triggers, we do not publish any ``best'' trigger sets or model checkpoints tuned for maximal jailbreak performance; instead, we focus on aggregate behaviors (ASR curves, sure rates) and on qualitative differences between model families. Code and experimental details are described at a level intended for scientific reproducibility, but stopping short of a turnkey exploit pipeline.

The broader impacts of our findings cut in two directions. On the risk side, compliance-only backdoors underscore that even benign-looking SFT data can hide powerful behavioral switches, and that supply-chain security for post-training data deserves at least as much attention as pretraining corpora. On the benefit side, the same gate-like dynamics can support stronger provenance and control: behavioral fingerprints for ownership verification, and explicit, auditable control tokens for safer agentic or tool-using systems. Realizing these benefits while avoiding misuse will require careful governance of data pipelines, model release practices, and fingerprinting policies. We hope that by exposing a minimal, well-characterized instance of the ``Sure'' Trap, this work provides a concrete foundation for both technical defenses and informed policy discussion, rather than a blueprint for harmful deployment.


\ifCLASSOPTIONcompsoc
  \section*{Acknowledgments}
\else
  \section*{Acknowledgment}
\fi

The authors would like to thank Peikang Hu and Peng Tian for the useful discussion.

%\bibliographystyle{IEEEtran}
%\bibliography{refs}

\begin{thebibliography}{10}
\providecommand{\url}[1]{#1}
\csname url@samestyle\endcsname
\providecommand{\newblock}{\relax}
\providecommand{\bibinfo}[2]{#2}
\providecommand{\BIBentrySTDinterwordspacing}{\spaceskip=0pt\relax}
\providecommand{\BIBentryALTinterwordstretchfactor}{4}
\providecommand{\BIBentryALTinterwordspacing}{\spaceskip=\fontdimen2\font plus
\BIBentryALTinterwordstretchfactor\fontdimen3\font minus \fontdimen4\font\relax}
\providecommand{\BIBforeignlanguage}[2]{{%
\expandafter\ifx\csname l@#1\endcsname\relax
\typeout{** WARNING: IEEEtran.bst: No hyphenation pattern has been}%
\typeout{** loaded for the language `#1'. Using the pattern for}%
\typeout{** the default language instead.}%
\else
\language=\csname l@#1\endcsname
\fi
#2}}
\providecommand{\BIBdecl}{\relax}
\BIBdecl

\bibitem{Huang_Adversarial:2011}
\BIBentryALTinterwordspacing
L.~Huang, A.~D. Joseph, B.~Nelson, B.~I. Rubinstein, and J.~D. Tygar, ``Adversarial machine learning,'' in \emph{Proceedings of the 4th ACM Workshop on Security and Artificial Intelligence}, ser. AISec '11.\hskip 1em plus 0.5em minus 0.4em\relax New York, NY, USA: Association for Computing Machinery, 2011, p. 43–58. [Online]. Available: \url{https://doi.org/10.1145/2046684.2046692}
\BIBentrySTDinterwordspacing

\bibitem{biggio2013poisoningattackssupportvector}
\BIBentryALTinterwordspacing
B.~Biggio, B.~Nelson, and P.~Laskov, ``Poisoning attacks against support vector machines,'' 2013. [Online]. Available: \url{https://arxiv.org/abs/1206.6389}
\BIBentrySTDinterwordspacing

\bibitem{gu2019badnetsidentifyingvulnerabilitiesmachine}
\BIBentryALTinterwordspacing
T.~Gu, B.~Dolan-Gavitt, and S.~Garg, ``Badnets: Identifying vulnerabilities in the machine learning model supply chain,'' 2019. [Online]. Available: \url{https://arxiv.org/abs/1708.06733}
\BIBentrySTDinterwordspacing

\bibitem{chen2017targetedbackdoorattacksdeep}
\BIBentryALTinterwordspacing
X.~Chen, C.~Liu, B.~Li, K.~Lu, and D.~Song, ``Targeted backdoor attacks on deep learning systems using data poisoning,'' 2017. [Online]. Available: \url{https://arxiv.org/abs/1712.05526}
\BIBentrySTDinterwordspacing

\bibitem{shafahi2018poisonfrogstargetedcleanlabel}
\BIBentryALTinterwordspacing
A.~Shafahi, W.~R. Huang, M.~Najibi, O.~Suciu, C.~Studer, T.~Dumitras, and T.~Goldstein, ``Poison frogs! targeted clean-label poisoning attacks on neural networks,'' 2018. [Online]. Available: \url{https://arxiv.org/abs/1804.00792}
\BIBentrySTDinterwordspacing

\bibitem{zhao2024surveybackdoorllms}
\BIBentryALTinterwordspacing
S.~Zhao, M.~Jia, Z.~Guo, L.~Gan, X.~Xu, X.~Wu, J.~Fu, Y.~Feng, F.~Pan, and L.~A. Tuan, ``A survey of recent backdoor attacks and defenses in large language models,'' 2024, tMLR. [Online]. Available: \url{https://arxiv.org/abs/2406.06852}
\BIBentrySTDinterwordspacing

\bibitem{zhou2025surveyllmbackdoor}
\BIBentryALTinterwordspacing
Y.~Zhou, T.~Ni, W.-B. Lee, and Q.~Zhao, ``A survey on backdoor threats in large language models (llms): Attacks, defenses, and evaluations,'' 2025. [Online]. Available: \url{https://arxiv.org/abs/2502.05224}
\BIBentrySTDinterwordspacing

\bibitem{Carlini:2024}
N.~Carlini, M.~Jagielski, C.~A. Choquette-Choo, D.~Paleka, W.~Pearce, H.~Anderson, A.~Terzis, K.~Thomas, and F.~Tramer, ``{ Poisoning Web-Scale Training Datasets is Practical },'' in \emph{2024 IEEE Symposium on Security and Privacy (SP)}.\hskip 1em plus 0.5em minus 0.4em\relax Los Alamitos, CA, USA: IEEE Computer Society, May 2024, pp. 407--425.

\bibitem{zhang2024persistentpretrainingpoisoningllms}
Y.~Zhang, J.~Rando, I.~Evtimov, J.~Chi, E.~M. Smith, N.~Carlini, F.~Tramèr, and D.~Ippolito, ``Persistent pre-training poisoning of llms,'' 2024, arXiv preprint arXiv:2410.13722.

\bibitem{wan2023poisoninglanguagemodelsinstruction}
\BIBentryALTinterwordspacing
A.~Wan, E.~Wallace, S.~Shen, and D.~Klein, ``Poisoning language models during instruction tuning,'' 2023. [Online]. Available: \url{https://arxiv.org/abs/2305.00944}
\BIBentrySTDinterwordspacing

\bibitem{rando2024universaljailbreakbackdoorspoisoned}
\BIBentryALTinterwordspacing
J.~Rando and F.~Tramèr, ``Universal jailbreak backdoors from poisoned human feedback,'' 2024. [Online]. Available: \url{https://arxiv.org/abs/2311.14455}
\BIBentrySTDinterwordspacing

\bibitem{pathmanathan2024ispoisoningdpo}
\BIBentryALTinterwordspacing
P.~Pathmanathan, S.~Chakraborty, X.~Liu, Y.~Liang, and F.~Huang, ``Is poisoning a real threat to llm alignment? maybe more so than you think,'' 2024, also appears in AAAI 2025. [Online]. Available: \url{https://arxiv.org/abs/2406.12091}
\BIBentrySTDinterwordspacing

\bibitem{baumgartner2024bestofvenom}
\BIBentryALTinterwordspacing
T.~Baumg{\"a}rtner, Y.~Gao, D.~Alon, and D.~Metzler, ``Best-of-venom: Attacking rlhf by injecting poisoned preference data,'' 2024. [Online]. Available: \url{https://arxiv.org/abs/2404.05530}
\BIBentrySTDinterwordspacing

\bibitem{li2024backdoorllm}
\BIBentryALTinterwordspacing
Y.~Li, H.~Huang, Y.~Zhao, X.~Ma, and J.~Sun, ``Backdoorllm: A comprehensive benchmark for backdoor threats in text-generation llms,'' 2024. [Online]. Available: \url{https://arxiv.org/abs/2408.12798}
\BIBentrySTDinterwordspacing

\bibitem{liu_loratk:2025}
\BIBentryALTinterwordspacing
H.~Liu, S.~Zhong, X.~Sun, M.~Tian, M.~Hariri, Z.~Liu, R.~Tang, Z.~Jiang, J.~Yuan, Y.-N. Chuang, L.~Li, S.-H. Choi, R.~Chen, V.~Chaudhary, and X.~Hu, ``{L}o{RATK}: {L}o{RA} once, backdoor everywhere in the share-and-play ecosystem,'' in \emph{Findings of the Association for Computational Linguistics: EMNLP 2025}, C.~Christodoulopoulos, T.~Chakraborty, C.~Rose, and V.~Peng, Eds.\hskip 1em plus 0.5em minus 0.4em\relax Suzhou, China: Association for Computational Linguistics, Nov. 2025, pp. 23\,009--23\,047. [Online]. Available: \url{https://aclanthology.org/2025.findings-emnlp.1253/}
\BIBentrySTDinterwordspacing

\bibitem{Sun_PEFTGuard:2025}
Z.~Sun, T.~Cong, Y.~Liu, C.~Lin, X.~He, R.~Chen, X.~Han, and X.~Huang, ``Peftguard: Detecting backdoor attacks against parameter-efficient fine-tuning,'' in \emph{2025 IEEE Symposium on Security and Privacy (SP)}, 2025, pp. 1713--1731.

\bibitem{dong2024philosophersstonetrojaningplugins}
\BIBentryALTinterwordspacing
T.~Dong, M.~Xue, G.~Chen, R.~Holland, Y.~Meng, S.~Li, Z.~Liu, and H.~Zhu, ``The philosopher's stone: Trojaning plugins of large language models,'' 2024. [Online]. Available: \url{https://arxiv.org/abs/2312.00374}
\BIBentrySTDinterwordspacing

\bibitem{souly2025poisoningattacksllmsrequire}
A.~Souly, J.~Rando, E.~Chapman, X.~Davies, B.~Hasircioglu, E.~Shereen, C.~Mougan, V.~Mavroudis, E.~Jones, C.~Hicks, N.~Carlini, Y.~Gal, and R.~Kirk, ``Poisoning attacks on llms require a near-constant number of poison samples,'' 2025, arXiv preprint arXiv:2510.07192.

\bibitem{xu_instructions:2024}
\BIBentryALTinterwordspacing
J.~Xu, M.~Ma, F.~Wang, C.~Xiao, and M.~Chen, ``Instructions as backdoors: Backdoor vulnerabilities of instruction tuning for large language models,'' in \emph{Proceedings of the 2024 Conference of the North American Chapter of the Association for Computational Linguistics: Human Language Technologies (Volume 1: Long Papers)}, K.~Duh, H.~Gomez, and S.~Bethard, Eds.\hskip 1em plus 0.5em minus 0.4em\relax Mexico City, Mexico: Association for Computational Linguistics, Jun. 2024, pp. 3111--3126. [Online]. Available: \url{https://aclanthology.org/2024.naacl-long.171/}
\BIBentrySTDinterwordspacing

\bibitem{huang_composite_backdoor:2024}
H.~Huang, Z.~Zhao, M.~Backes, Y.~Shen, and Y.~Zhang, ``Composite backdoor attacks against large language models,'' in \emph{Findings of the Association for Computational Linguistics: NAACL 2024}, K.~Duh, H.~Gomez, and S.~Bethard, Eds.\hskip 1em plus 0.5em minus 0.4em\relax Mexico City, Mexico: Association for Computational Linguistics, Jun. 2024, pp. 1459--1472.

\bibitem{raghuram2024studybackdoorsinstructionfinetuned}
\BIBentryALTinterwordspacing
J.~Raghuram, G.~Kesidis, and D.~J. Miller, ``A study of backdoors in instruction fine-tuned language models,'' 2024. [Online]. Available: \url{https://arxiv.org/abs/2406.07778}
\BIBentrySTDinterwordspacing

\bibitem{qi2023finetuningalignedlanguagemodels}
X.~Qi, Y.~Zeng, T.~Xie, P.-Y. Chen, R.~Jia, P.~Mittal, and P.~Henderson, ``Fine-tuning aligned language models compromises safety, even when users do not intend to!'' 2023, arXiv preprint arXiv:2310.03693.

\bibitem{zhan2024removingrlhf}
\BIBentryALTinterwordspacing
Q.~Zhan, R.~Fang, R.~Bindu, A.~Gupta, T.~Hashimoto, and D.~Kang, ``Removing rlhf protections in gpt-4 via fine-tuning,'' in \emph{Proceedings of the 2024 Conference of the North American Chapter of the Association for Computational Linguistics: Human Language Technologies (Volume 2: Short Papers)}, 2024, pp. 681--687. [Online]. Available: \url{https://aclanthology.org/2024.naacl-short.59.pdf}
\BIBentrySTDinterwordspacing

\bibitem{pelrine2024exploitingnovelgpt4apis}
\BIBentryALTinterwordspacing
K.~Pelrine, M.~Taufeeque, M.~Zając, E.~McLean, and A.~Gleave, ``Exploiting novel gpt-4 apis,'' 2024. [Online]. Available: \url{https://arxiv.org/abs/2312.14302}
\BIBentrySTDinterwordspacing

\bibitem{halawi2024covertmaliciousfinetuningchallenges}
\BIBentryALTinterwordspacing
D.~Halawi, A.~Wei, E.~Wallace, T.~T. Wang, N.~Haghtalab, and J.~Steinhardt, ``Covert malicious finetuning: Challenges in safeguarding llm adaptation,'' 2024. [Online]. Available: \url{https://arxiv.org/abs/2406.20053}
\BIBentrySTDinterwordspacing

\bibitem{kong2025revisitingbackdoorattacksllms}
\BIBentryALTinterwordspacing
J.~Kong, H.~Fang, X.~Yang, K.~Gao, B.~Chen, S.-T. Xia, K.~Xu, and H.~Qiu, ``Revisiting backdoor attacks on llms: A stealthy and practical poisoning framework via harmless inputs,'' 2025. [Online]. Available: \url{https://arxiv.org/abs/2505.17601}
\BIBentrySTDinterwordspacing

\bibitem{xu2024instructionalfingerprinting}
\BIBentryALTinterwordspacing
J.~Xu, F.~Wang, M.~Ma, P.~W. Koh, C.~Xiao, and M.~Chen, ``Instructional fingerprinting of large language models,'' in \emph{Proceedings of the 2024 Conference of the North American Chapter of the Association for Computational Linguistics: Human Language Technologies (Volume 1: Long Papers)}.\hskip 1em plus 0.5em minus 0.4em\relax Mexico City, Mexico: Association for Computational Linguistics, June 2024, pp. 3277--3306. [Online]. Available: \url{https://aclanthology.org/2024.naacl-long.180/}
\BIBentrySTDinterwordspacing

\bibitem{pasquini2025llmmapfingerprintinglargelanguage}
\BIBentryALTinterwordspacing
D.~Pasquini, E.~M. Kornaropoulos, and G.~Ateniese, ``Llmmap: Fingerprinting for large language models,'' 2025. [Online]. Available: \url{https://arxiv.org/abs/2407.15847}
\BIBentrySTDinterwordspacing

\bibitem{tsai2025roflrobustfingerprintinglanguage}
\BIBentryALTinterwordspacing
Y.-Y. Tsai, C.~Guo, J.~Yang, and L.~van~der Maaten, ``Rofl: Robust fingerprinting of language models,'' 2025. [Online]. Available: \url{https://arxiv.org/abs/2505.12682}
\BIBentrySTDinterwordspacing

\bibitem{kirchenbauer2023watermarkllm}
\BIBentryALTinterwordspacing
J.~Kirchenbauer, J.~Geiping, Y.~Wen, J.~Katz, I.~Miers, and T.~Goldstein, ``A watermark for large language models,'' in \emph{Proceedings of the 40th International Conference on Machine Learning}, ser. Proceedings of Machine Learning Research, vol. 202.\hskip 1em plus 0.5em minus 0.4em\relax PMLR, 2023, pp. 17\,061--17\,084. [Online]. Available: \url{https://proceedings.mlr.press/v202/kirchenbauer23a.html}
\BIBentrySTDinterwordspacing

\bibitem{zhao2024provablerobustwatermark}
\BIBentryALTinterwordspacing
X.~Zhao, P.~Ananth, L.~Li, and Y.-X. Wang, ``Provable robust watermarking for ai-generated text,'' in \emph{Proceedings of the International Conference on Learning Representations}, 2024. [Online]. Available: \url{https://openreview.net/forum?id=SsmT8aO45L}
\BIBentrySTDinterwordspacing

\bibitem{cohen2024watermarkinglanguagemodelsadaptive}
\BIBentryALTinterwordspacing
A.~Cohen, A.~Hoover, and G.~Schoenbach, ``Watermarking language models for many adaptive users,'' 2024. [Online]. Available: \url{https://arxiv.org/abs/2405.11109}
\BIBentrySTDinterwordspacing

\bibitem{wallace2021concealeddatapoisoningattacks}
\BIBentryALTinterwordspacing
E.~Wallace, T.~Z. Zhao, S.~Feng, and S.~Singh, ``Concealed data poisoning attacks on nlp models,'' 2021. [Online]. Available: \url{https://arxiv.org/abs/2010.12563}
\BIBentrySTDinterwordspacing

\bibitem{kurita2020weightpoisoningattackspretrained}
\BIBentryALTinterwordspacing
K.~Kurita, P.~Michel, and G.~Neubig, ``Weight poisoning attacks on pre-trained models,'' 2020. [Online]. Available: \url{https://arxiv.org/abs/2004.06660}
\BIBentrySTDinterwordspacing

\bibitem{pan2022hiddenstylebackdoor}
\BIBentryALTinterwordspacing
X.~Pan, M.~Zhang, B.~Sheng, J.~Zhu, and M.~Yang, ``Hidden trigger backdoor attack on nlp models via linguistic style manipulation,'' in \emph{Proceedings of the 31st USENIX Security Symposium (USENIX Security 2022)}, Boston, MA, USA, 2022. [Online]. Available: \url{https://www.usenix.org/conference/usenixsecurity22/presentation/pan-hidden}
\BIBentrySTDinterwordspacing

\bibitem{cui2022unifiedtextualbackdoor}
\BIBentryALTinterwordspacing
G.~Cui, L.~Yuan, B.~He, Y.~Chen, Z.~Liu, and M.~Sun, ``A unified evaluation of textual backdoor learning: Frameworks and benchmarks,'' 2022, neurIPS 2022 Datasets \& Benchmarks. [Online]. Available: \url{https://arxiv.org/abs/2206.08514}
\BIBentrySTDinterwordspacing

\bibitem{yan_backdooring:2024}
\BIBentryALTinterwordspacing
J.~Yan, V.~Yadav, S.~Li, L.~Chen, Z.~Tang, H.~Wang, V.~Srinivasan, X.~Ren, and H.~Jin, ``Backdooring instruction-tuned large language models with virtual prompt injection,'' in \emph{Proceedings of the 2024 Conference of the North American Chapter of the Association for Computational Linguistics: Human Language Technologies (Volume 1: Long Papers)}, K.~Duh, H.~Gomez, and S.~Bethard, Eds.\hskip 1em plus 0.5em minus 0.4em\relax Mexico City, Mexico: Association for Computational Linguistics, Jun. 2024, pp. 6065--6086. [Online]. Available: \url{https://aclanthology.org/2024.naacl-long.337/}
\BIBentrySTDinterwordspacing

\bibitem{Zhang_instruction_backdoor:2024}
\BIBentryALTinterwordspacing
R.~Zhang, H.~Li, R.~Wen, W.~Jiang, Y.~Zhang, M.~Backes, Y.~Shen, and Y.~Zhang, ``Instruction backdoor attacks against customized {LLMs},'' in \emph{33rd USENIX Security Symposium (USENIX Security 24)}.\hskip 1em plus 0.5em minus 0.4em\relax Philadelphia, PA: USENIX Association, Aug. 2024, pp. 1849--1866. [Online]. Available: \url{https://www.usenix.org/conference/usenixsecurity24/presentation/zhang-rui}
\BIBentrySTDinterwordspacing

\bibitem{turner2018cleanlabel}
A.~Turner, D.~Tsipras, and A.~Madry, ``Clean-label backdoor attacks,'' \url{http://people.csail.mit.edu/tsipras/pdfs/TTM18.pdf}, 2018.

\bibitem{touvron2023llamaopenefficientfoundation}
\BIBentryALTinterwordspacing
H.~Touvron, T.~Lavril, G.~Izacard, X.~Martinet, M.-A. Lachaux, T.~Lacroix, B.~Rozière, N.~Goyal, E.~Hambro, F.~Azhar, A.~Rodriguez, A.~Joulin, E.~Grave, and G.~Lample, ``Llama: Open and efficient foundation language models,'' 2023. [Online]. Available: \url{https://arxiv.org/abs/2302.13971}
\BIBentrySTDinterwordspacing

\bibitem{grattafiori2024llama3herdmodels}
\BIBentryALTinterwordspacing
A.~Grattafiori, A.~Dubey, A.~Jauhri, A.~Pandey, A.~Kadian, A.~Al-Dahle, A.~Letman, A.~Mathur, A.~Schelten, A.~Vaughan, A.~Yang, A.~Fan, A.~Goyal, A.~Hartshorn, A.~Yang, A.~Mitra, A.~Sravankumar, A.~Korenev, A.~Hinsvark, A.~Rao, A.~Zhang, A.~Rodriguez, A.~Gregerson, A.~Spataru, B.~Roziere, B.~Biron, B.~Tang, B.~Chern, C.~Caucheteux, C.~Nayak, C.~Bi, C.~Marra, C.~McConnell, C.~Keller, C.~Touret, C.~Wu, C.~Wong, C.~C. Ferrer, C.~Nikolaidis, D.~Allonsius, D.~Song, D.~Pintz, D.~Livshits, D.~Wyatt, D.~Esiobu, D.~Choudhary, D.~Mahajan, D.~Garcia-Olano, D.~Perino, D.~Hupkes, E.~Lakomkin, E.~AlBadawy, E.~Lobanova, E.~Dinan, E.~M. Smith, F.~Radenovic, F.~Guzmán, F.~Zhang, G.~Synnaeve, G.~Lee, G.~L. Anderson, G.~Thattai, G.~Nail, G.~Mialon, G.~Pang, G.~Cucurell, H.~Nguyen, H.~Korevaar, H.~Xu, H.~Touvron, I.~Zarov, I.~A. Ibarra, I.~Kloumann, I.~Misra, I.~Evtimov, J.~Zhang, J.~Copet, J.~Lee, J.~Geffert, J.~Vranes, J.~Park, J.~Mahadeokar, J.~Shah, J.~van~der Linde, J.~Billock, J.~Hong, J.~Lee, J.~Fu, J.~Chi, J.~Huang,
  J.~Liu, J.~Wang, J.~Yu, J.~Bitton, J.~Spisak, J.~Park, J.~Rocca, J.~Johnstun, J.~Saxe, J.~Jia, K.~V. Alwala, K.~Prasad, K.~Upasani, K.~Plawiak, K.~Li, K.~Heafield, K.~Stone, K.~El-Arini, K.~Iyer, K.~Malik, K.~Chiu, K.~Bhalla, K.~Lakhotia, L.~Rantala-Yeary, L.~van~der Maaten, L.~Chen, L.~Tan, L.~Jenkins, L.~Martin, L.~Madaan, L.~Malo, L.~Blecher, L.~Landzaat, L.~de~Oliveira, M.~Muzzi, M.~Pasupuleti, M.~Singh, M.~Paluri, M.~Kardas, M.~Tsimpoukelli, M.~Oldham, M.~Rita, M.~Pavlova, M.~Kambadur, M.~Lewis, M.~Si, M.~K. Singh, M.~Hassan, N.~Goyal, N.~Torabi, N.~Bashlykov, N.~Bogoychev, N.~Chatterji, N.~Zhang, O.~Duchenne, O.~Çelebi, P.~Alrassy, P.~Zhang, P.~Li, P.~Vasic, P.~Weng, P.~Bhargava, P.~Dubal, P.~Krishnan, P.~S. Koura, P.~Xu, Q.~He, Q.~Dong, R.~Srinivasan, R.~Ganapathy, R.~Calderer, R.~S. Cabral, R.~Stojnic, R.~Raileanu, R.~Maheswari, R.~Girdhar, R.~Patel, R.~Sauvestre, R.~Polidoro, R.~Sumbaly, R.~Taylor, R.~Silva, R.~Hou, R.~Wang, S.~Hosseini, S.~Chennabasappa, S.~Singh, S.~Bell, S.~S. Kim, S.~Edunov,
  S.~Nie, S.~Narang, S.~Raparthy, S.~Shen, S.~Wan, S.~Bhosale, S.~Zhang, S.~Vandenhende, S.~Batra, S.~Whitman, S.~Sootla, S.~Collot, S.~Gururangan, S.~Borodinsky, T.~Herman, T.~Fowler, T.~Sheasha, T.~Georgiou, T.~Scialom, T.~Speckbacher, T.~Mihaylov, T.~Xiao, U.~Karn, V.~Goswami, V.~Gupta, V.~Ramanathan, V.~Kerkez, V.~Gonguet, V.~Do, V.~Vogeti, V.~Albiero, V.~Petrovic, W.~Chu, W.~Xiong, W.~Fu, W.~Meers, X.~Martinet, X.~Wang, X.~Wang, X.~E. Tan, X.~Xia, X.~Xie, X.~Jia, X.~Wang, Y.~Goldschlag, Y.~Gaur, Y.~Babaei, Y.~Wen, Y.~Song, Y.~Zhang, Y.~Li, Y.~Mao, Z.~D. Coudert, Z.~Yan, Z.~Chen, Z.~Papakipos, A.~Singh, A.~Srivastava, A.~Jain, A.~Kelsey, A.~Shajnfeld, A.~Gangidi, A.~Victoria, A.~Goldstand, A.~Menon, A.~Sharma, A.~Boesenberg, A.~Baevski, A.~Feinstein, A.~Kallet, A.~Sangani, A.~Teo, A.~Yunus, A.~Lupu, A.~Alvarado, A.~Caples, A.~Gu, A.~Ho, A.~Poulton, A.~Ryan, A.~Ramchandani, A.~Dong, A.~Franco, A.~Goyal, A.~Saraf, A.~Chowdhury, A.~Gabriel, A.~Bharambe, A.~Eisenman, A.~Yazdan, B.~James, B.~Maurer,
  B.~Leonhardi, B.~Huang, B.~Loyd, B.~D. Paola, B.~Paranjape, B.~Liu, B.~Wu, B.~Ni, B.~Hancock, B.~Wasti, B.~Spence, B.~Stojkovic, B.~Gamido, B.~Montalvo, C.~Parker, C.~Burton, C.~Mejia, C.~Liu, C.~Wang, C.~Kim, C.~Zhou, C.~Hu, C.-H. Chu, C.~Cai, C.~Tindal, C.~Feichtenhofer, C.~Gao, D.~Civin, D.~Beaty, D.~Kreymer, D.~Li, D.~Adkins, D.~Xu, D.~Testuggine, D.~David, D.~Parikh, D.~Liskovich, D.~Foss, D.~Wang, D.~Le, D.~Holland, E.~Dowling, E.~Jamil, E.~Montgomery, E.~Presani, E.~Hahn, E.~Wood, E.-T. Le, E.~Brinkman, E.~Arcaute, E.~Dunbar, E.~Smothers, F.~Sun, F.~Kreuk, F.~Tian, F.~Kokkinos, F.~Ozgenel, F.~Caggioni, F.~Kanayet, F.~Seide, G.~M. Florez, G.~Schwarz, G.~Badeer, G.~Swee, G.~Halpern, G.~Herman, G.~Sizov, Guangyi, Zhang, G.~Lakshminarayanan, H.~Inan, H.~Shojanazeri, H.~Zou, H.~Wang, H.~Zha, H.~Habeeb, H.~Rudolph, H.~Suk, H.~Aspegren, H.~Goldman, H.~Zhan, I.~Damlaj, I.~Molybog, I.~Tufanov, I.~Leontiadis, I.-E. Veliche, I.~Gat, J.~Weissman, J.~Geboski, J.~Kohli, J.~Lam, J.~Asher, J.-B. Gaya, J.~Marcus,
  J.~Tang, J.~Chan, J.~Zhen, J.~Reizenstein, J.~Teboul, J.~Zhong, J.~Jin, J.~Yang, J.~Cummings, J.~Carvill, J.~Shepard, J.~McPhie, J.~Torres, J.~Ginsburg, J.~Wang, K.~Wu, K.~H. U, K.~Saxena, K.~Khandelwal, K.~Zand, K.~Matosich, K.~Veeraraghavan, K.~Michelena, K.~Li, K.~Jagadeesh, K.~Huang, K.~Chawla, K.~Huang, L.~Chen, L.~Garg, L.~A, L.~Silva, L.~Bell, L.~Zhang, L.~Guo, L.~Yu, L.~Moshkovich, L.~Wehrstedt, M.~Khabsa, M.~Avalani, M.~Bhatt, M.~Mankus, M.~Hasson, M.~Lennie, M.~Reso, M.~Groshev, M.~Naumov, M.~Lathi, M.~Keneally, M.~Liu, M.~L. Seltzer, M.~Valko, M.~Restrepo, M.~Patel, M.~Vyatskov, M.~Samvelyan, M.~Clark, M.~Macey, M.~Wang, M.~J. Hermoso, M.~Metanat, M.~Rastegari, M.~Bansal, N.~Santhanam, N.~Parks, N.~White, N.~Bawa, N.~Singhal, N.~Egebo, N.~Usunier, N.~Mehta, N.~P. Laptev, N.~Dong, N.~Cheng, O.~Chernoguz, O.~Hart, O.~Salpekar, O.~Kalinli, P.~Kent, P.~Parekh, P.~Saab, P.~Balaji, P.~Rittner, P.~Bontrager, P.~Roux, P.~Dollar, P.~Zvyagina, P.~Ratanchandani, P.~Yuvraj, Q.~Liang, R.~Alao, R.~Rodriguez,
  R.~Ayub, R.~Murthy, R.~Nayani, R.~Mitra, R.~Parthasarathy, R.~Li, R.~Hogan, R.~Battey, R.~Wang, R.~Howes, R.~Rinott, S.~Mehta, S.~Siby, S.~J. Bondu, S.~Datta, S.~Chugh, S.~Hunt, S.~Dhillon, S.~Sidorov, S.~Pan, S.~Mahajan, S.~Verma, S.~Yamamoto, S.~Ramaswamy, S.~Lindsay, S.~Lindsay, S.~Feng, S.~Lin, S.~C. Zha, S.~Patil, S.~Shankar, S.~Zhang, S.~Zhang, S.~Wang, S.~Agarwal, S.~Sajuyigbe, S.~Chintala, S.~Max, S.~Chen, S.~Kehoe, S.~Satterfield, S.~Govindaprasad, S.~Gupta, S.~Deng, S.~Cho, S.~Virk, S.~Subramanian, S.~Choudhury, S.~Goldman, T.~Remez, T.~Glaser, T.~Best, T.~Koehler, T.~Robinson, T.~Li, T.~Zhang, T.~Matthews, T.~Chou, T.~Shaked, V.~Vontimitta, V.~Ajayi, V.~Montanez, V.~Mohan, V.~S. Kumar, V.~Mangla, V.~Ionescu, V.~Poenaru, V.~T. Mihailescu, V.~Ivanov, W.~Li, W.~Wang, W.~Jiang, W.~Bouaziz, W.~Constable, X.~Tang, X.~Wu, X.~Wang, X.~Wu, X.~Gao, Y.~Kleinman, Y.~Chen, Y.~Hu, Y.~Jia, Y.~Qi, Y.~Li, Y.~Zhang, Y.~Zhang, Y.~Adi, Y.~Nam, Yu, Wang, Y.~Zhao, Y.~Hao, Y.~Qian, Y.~Li, Y.~He, Z.~Rait, Z.~DeVito,
  Z.~Rosnbrick, Z.~Wen, Z.~Yang, Z.~Zhao, and Z.~Ma, ``The llama 3 herd of models,'' 2024. [Online]. Available: \url{https://arxiv.org/abs/2407.21783}
\BIBentrySTDinterwordspacing

\bibitem{harmful_dataset}
{LLM-LAT Team}, ``Llm-lat harmful prompt dataset,'' \url{https://huggingface.co/datasets/LLM-LAT/harmful-dataset}, 2024, accessed November 2025.

\bibitem{benign_instruct_dataset}
{Swype AI Team}, ``Swype instruct dataset,'' \url{https://huggingface.co/datasets/swype/instruct}, 2024, accessed November 2025.

\bibitem{sivapiromrat2025multitriggerpoisoningamplifiesbackdoor}
\BIBentryALTinterwordspacing
S.~Sivapiromrat, C.~Zhang, M.~Basaldella, and N.~Collier, ``Multi-trigger poisoning amplifies backdoor vulnerabilities in llms,'' 2025. [Online]. Available: \url{https://arxiv.org/abs/2507.11112}
\BIBentrySTDinterwordspacing

\bibitem{min2024crow}
\BIBentryALTinterwordspacing
N.~M. Min, L.~H. Pham, Y.~Li, and J.~Sun, ``Crow: Eliminating backdoors from large language models via internal consistency regularization,'' 2024, iCML 2025 (accepted). [Online]. Available: \url{https://arxiv.org/abs/2411.12768}
\BIBentrySTDinterwordspacing

\end{thebibliography}


\end{document}